\def\year{2021}\relax
\documentclass[letterpaper]{article} 
\usepackage{aaai21}  
\usepackage{times}  
\usepackage{helvet} 
\usepackage{courier}  
\usepackage[hyphens]{url}  
\usepackage{graphicx} 
\usepackage{graphicx}  
\usepackage{multirow}
\usepackage{amsmath}
\usepackage{booktabs}
\usepackage{subfigure}
\usepackage{color}
\usepackage{lipsum}
\usepackage{natbib}  
\usepackage{caption} 
\title{Weakly Supervised Semantic Segmentation for Large-Scale Point Cloud}
\author {
    Yachao Zhang\textsuperscript{\rm 1}, 
    Zonghao Li\textsuperscript{\rm 1}, 
    Yuan Xie\textsuperscript{\rm 2 \footnote{Corresponding Author} }, 
    Yanyun Qu\textsuperscript{\rm 1 $^*$}, 
    Cuihua Li\textsuperscript{\rm 1}, 
    Tao Mei\textsuperscript{\rm 3} \\
}
\affiliations {
    \textsuperscript{\rm 1} School of Informatics, Xiamen University, Fujian, China\\
    \textsuperscript{\rm 2} School of Computer Science and Technology, East China Normal University, Shanghai, China \\
    \textsuperscript{\rm 3} JD AI Research, Beijing, China \\
     \{yachaozhang,zonghaoli\}@stu.xmu.edu.cn, yxie@cs.ecnu.edu.cn,
    \{yyqu,chli\}@xmu.edu.cn, tmei@jd.com
}
\begin{document}

\maketitle

\begin{abstract}
	Existing methods for large-scale point cloud semantic segmentation require expensive, tedious and error-prone manual point-wise annotations. Intuitively, weakly supervised training is a direct solution to reduce the cost of labeling. However, for weakly supervised large-scale point cloud semantic segmentation, too few annotations will inevitably lead to ineffective learning of network. We propose an effective weakly supervised method containing two components to solve the above problem. Firstly, we construct a pretext task, \textit{i.e.,} point cloud colorization, with a self-supervised learning to transfer the learned prior knowledge from a large amount of unlabeled point cloud to a weakly supervised network. In this way, the representation capability of the weakly supervised network can be improved by the guidance from a heterogeneous task. Besides, to generate pseudo label for unlabeled data, a sparse label propagation mechanism is proposed with the help of generated class prototypes, which is used to measure the classification confidence of unlabeled point. Our method is evaluated on large-scale point cloud datasets with different scenarios including indoor and outdoor. The experimental results show the large gain against existing weakly supervised and comparable results to fully supervised methods\footnote{Code based on mindspore: https://github.com/dmcv-ecnu/MindSpore\_ModelZoo/tree/main/WS3\_MindSpore}.
	
\end{abstract}

\section{Introduction}

3D scene understanding is required in numerous applications, in particular robotics, autonomous driving and virtual reality. Large-scale point cloud semantic segmentation as a fundamental task attracts more and more attention. The success of deep neural networks in point cloud semantic segmentation is attribute to their ability to scale up with more well-labeled training data \cite{qi2017pointnet,qi2017pointnet++,shellnet,li2018so,DGCNN,wu2019pointconv,yang2019modeling,RandLA-Net}. 

Fully supervised point cloud semantic segmentation methods need expensive, tedious and error-prone manual point-wise annotations. One direct solution is to achieve effective segmentation via weakly supervised learning with annotating partial points or semantic category which is still in its infancy. Xu and Lee proposed a weakly supervised approach \cite{xu2020weakly} whose results are close to the previous fully supervised performance with $10\times$ fewer labeled points. This approach only deal with an instance (ShapeNetPart) or a block ($1m \times 1m$ on S3DIS) of point cloud in a small scale. 
\begin{figure}[t]
	\centering
	\includegraphics[width=8.25cm]{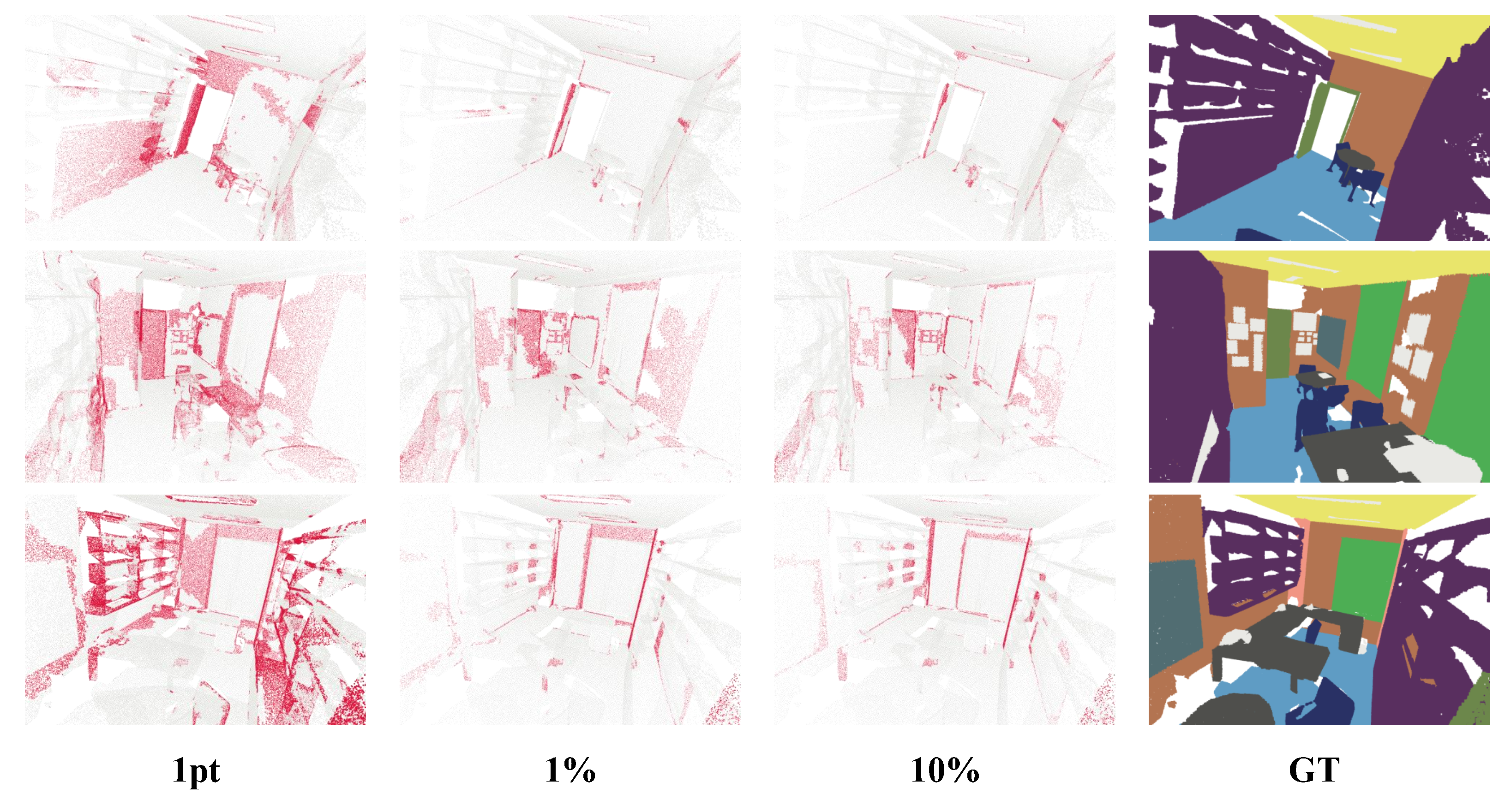}
	\caption{Visualize the results of semantic segmentation, the misclassification points are signed in red. From left to right, the columns show the results with one labeled point, 1\% labeled points, 10\% labeled points and the ground truth.}
	\label{vis_ss}
\end{figure}

For large-scale point clouds, the network cannot effectively learn feature representation with a few labeled points. We consider the two strategies to solve this problem: 1) What prior knowledge can be transferred to the segmentation network for improving feature representation in weakly supervised learning? 2) How are the labels of given points propagated to unlabeled points effectively?

Firstly, transfer learning provides a promising way and is relatively successful in 2D vision tasks, such as image classification, segmentation, and so on. But point cloud tasks lack a well-annotated and category-extensive dataset like ImageNet \cite{imagenet} for pre-training. We consider to use self-supervised learning for knowledge transfer. However, how to use enormous amounts of unlabeled point cloud to generate labels by itself and learn a semantically related representation for subsequent tasks is very challenging. Secondly, propagating pseudo labels to unlabeled points is a common method for weakly supervised and semi-supervised task to learn effectively. Usually, this requires constructing a fully-connected graph with all points. However, for large-scale point clouds ($\sim 10^6$ points), the fully-connected graph is unachievable due to large GPU memory consumption. 

To address the above difficulties, we present an effective weakly supervised method for large-scale point cloud semantic segmentation. Firstly, we notice that points of the same semantic class have similar color distribution, and point cloud with color is essentially free. We choose point cloud colorization as a pretext task for transfer learning. Specifically, we use color space transformation to construct a self-supervised network for color space completion. We further introduce a local perceptual regularization term to enhance the local representation which is consistent with the goal of segmentation task. As a result, it can learn a prior-based initialization distribution. After that, we use the pre-trained knowledge to fine-tune weakly semantic segmentation network. Our learning scheme allows to transfer the knowledge from the self-supervised task to the weakly supervised task and improves the effectiveness of feature representation. Moreover, we propose a sparse label propagation method induced by class prototype which can gradually propagate pseudo labels to unlabeled points. It is computational friendly that can be adapted to large-scale tasks.

Our contributions can be summarized as follows:
\begin{itemize}	
	
	\item A weakly supervised segmentation method is proposed for large-scale point cloud which only needs very small number of labeled points.
	
	\item We adopt the heterogeneous transfer learning method and construct a self-supervised pretext task by point cloud colorization. It can learn a prior distribution and be generalized well to segmentation tasks.
	
	\item We propose an efficient sparse label propagation method which can propagate labels to unlabeled points. It expands the supervision information and has low computational complexity. 
	
	\item Extensive experimental results demonstrate that our method achieves the comparable to or even exceeds fully supervised competitors.
	
\end{itemize}

\section{Related Work}
\textbf{Semantic segmentation for large-scale point cloud.}
PointNet and PointNet++ \cite{qi2017pointnet,qi2017pointnet++} are pioneering approaches for point clouds. While recent works have shown promising results on small point clouds, most of them cannot directly scale up to large-scale point clouds ($\sim 10^6$ points) due to high computational and memory costs \cite{RandLA-Net}. 

SPG \cite{landrieu2018large} processes the large-scale point clouds through a graph convolution-based method. FCPN \cite{rethage2018fully} preprocesses large-scale point clouds into voxels. However, both the graph partitioning and voxelization are computationally expensive. Recently, RandLA-Net \cite{RandLA-Net} provides an efficient and lightweight neural architecture for large-scale point cloud semantic segmentation. The state-of-the-art methods mentioned above are fully supervised that require a large number of point clouds with dense point-wise annotation. Obviously, this point-wise annotation is labor-intensive and time-consuming.

\textbf{Self-supervised learning on point cloud.}
Self-supervised learning springs up in computer vision recently. It aims to learn good representations from unlabeled visual data, reducing or even eliminating the need for the costly collection of manual labels \cite{newell2020useful}. Recent self-supervised learning makes great successes in feature representation which achieves comparable or outperforms results produced by supervised pre-training \cite{bachman2019learning,he2020momentum}. 

Unlike 2D images, self-supervised learning in point cloud is rarely used. Previous works on unsupervised 3D representation learning \cite{achlioptas2018learning,gadelha2018multiresolution,hassani2019unsupervised,li2018so,sauder2019self,yang2018foldingnet} mainly focuse on representing an instance (ShapeNet \cite{chang2015shapenet}). It is difficult to directly apply the feature representation learned on an instance to real-word large-scale point cloud tasks due to the large domain gaps \cite{pointcontrast}. 

PointContrast \cite{pointcontrast} concerns on contrastive embeddings and proposed a pre-training task for 3D point cloud understanding. The core of this method is that different views of point cloud should be mapped to similar embeddings for matched points. It achieves promising results on fully-supervised downstream tasks. Xu and Lee \cite{xu2020weakly} proposed a Siamese self-supervision branch by augmenting the training sample with a random in-plane rotation and flipping, and then made the original and augmented point-wise predictions be consistent. This method \cite{xu2020weakly} treats self-supervision as a branch of multi-task and only the current training samples are used. Therefore, it cannot make full use of other massive point clouds to learn representations with strong generalization.

\textbf{Weakly Supervised Point Cloud Semantic Segmentation.}
There are few researches on weakly supervised point cloud semantic segmentation. Following the weakly supervised manner called incomplete supervision in \cite{zhou2018brief}, the recent work \cite{xu2020weakly} utilizes $10 \times$ fewer labeled points to achieve comparable performance to fully supervised method in small scale part segmentation and small blocks ( $\sim 10^3$ points) semantic segmentation. MPRM \cite{wei2020multi} introduces a multi-path region mining module to generate pseudo point-level label from a classification network. The classification network is trained by the subcloud-level weakly labels. For the scene-level as an input, the performance will degrade. However, up to now, none of them can be generalized to large-scale problem well.

\section{Proposed Method}

\subsection{Overview}
\begin{figure*}[t]
	\centering
	\includegraphics[width=16.5cm]{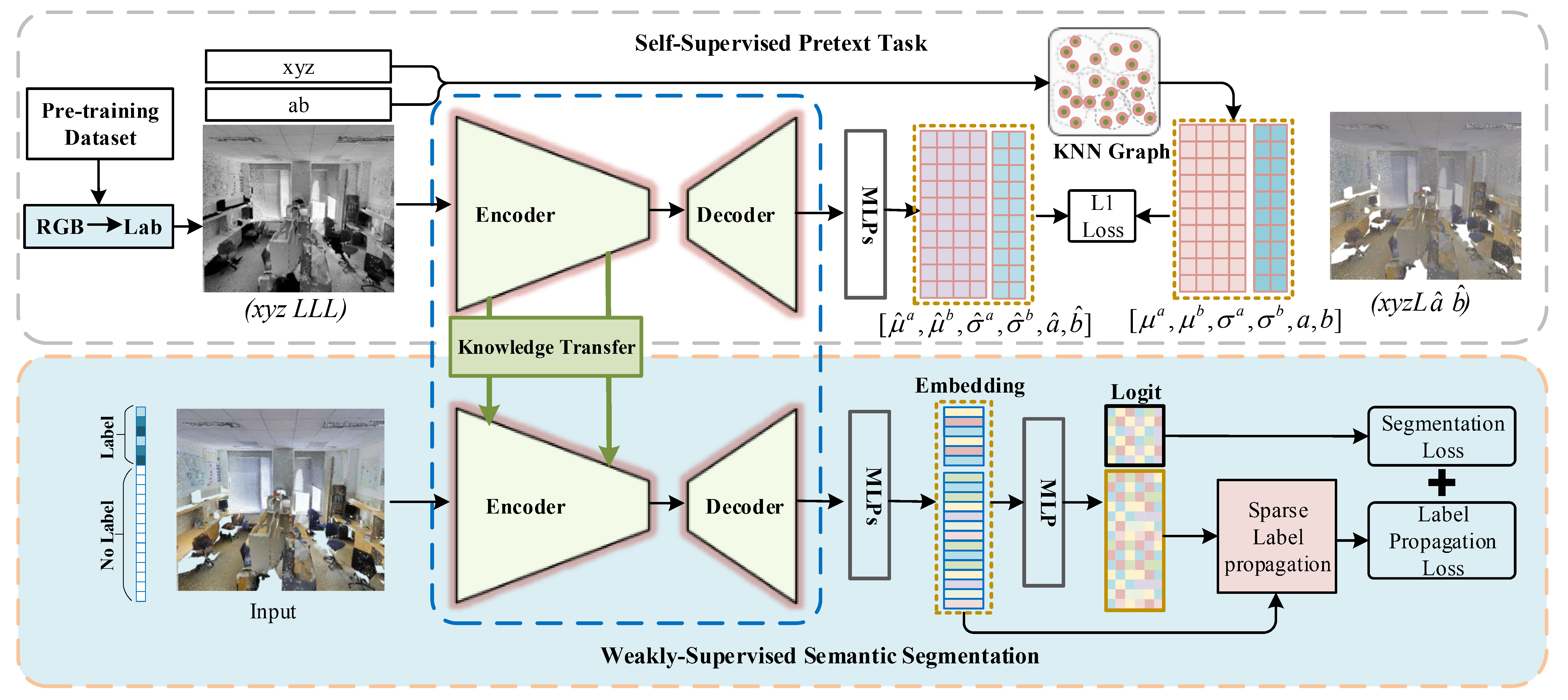}
	\caption{The framework of our method consists of three parts: i) Self-supervised pretext task learns a prior knowledge. ii) The prior knowledge is used to fine-tune the weakly-supervised semantic segmentation network. iii) Sparse label propagation generates pseudo label for unlabeled data to improves the effectiveness of weakly-supervised task.}
	\label{pipeline}
\end{figure*}

Our method aims to exploit the knowledge transfer and label propagation to solve the problem of unstable and poor representation produced by the network under weakly supervised large-scale point cloud segmentation. We propose an effective weakly supervised large-scale method and depict the overview framework in Figure \ref{pipeline}.

We take point cloud colorization as a self-supervised pretext task to learn a prior-based initialization distribution. A local perceptual regularization is proposed to learn the contextual information. Then, we use pre-trained parameters of encoder to initialize the weakly-supervised network for improving the effectiveness of feature presentation.

Furthermore, we leverage labeled points to directly supervise the network and fine-tune network parameters. We also introduce a non-parametric label propagation method for weakly supervised semantic segmentation. Some unlabeled points are assigned pseudo labels through the similarity between class prototypes \cite{li2020unsupervised,qiu2017learning} and embeddings of unlabeled points. Therefore, more supervised information is introduced to improve effectiveness of training. Considering of computational and memory efficiency for large-scale point cloud, we choose RandLA-Net \cite{RandLA-Net} as the backbone, which is an efficient and lightweight neural architecture for large-scale point clouds semantic segmentation. In the following, we describe the self-supervised pretext task and the sparse label propagation method.

\subsection{Self-Supervised Pretext Task}

Colorization provides a powerful supervisory signal unlike training from scratch in 2D vision task. Training data are easy to collect, so any point cloud with color can be used as training data. Due to the progress of point cloud acquisition equipment, we have access to enormous amounts of unlabeled point cloud data with color information. We investigate and implement self-supervised learning on point cloud colorization which is treated as a pretext task. Point colorization aims to guide the self-supervised model to learn the feature representation. 

It is recognized that the $\mathbf{Lab}$ color space favor for perceptual distance \cite{zhang2016colorful,larsson2017colorization}. Thus, we perform point cloud colorization by $\textbf{a}, \textbf{b}$ completion in this colorspace. Given the lightness channel ${\textbf{L}}$, the network predicts the $\textbf{a}$ and $\textbf{b}$ color channels and the local Gaussian distribution for each point. Notably, the value in channel ${L}$ is replicated to three times of each point to keep the same dimension as the input of segmentation task. Therefore, the input point cloud  $ X^{s}=[x_1,x_2,...,x_{N^{s}}] \in \mathcal{R}^{N^{s}\times 6}$ consists of $N^{s}$ 3D points with the \textit{xyz} coordinates and three $\textbf{L}$. $N^{s}$ is the number of points in one point cloud.

Moreover, we implement RandLA-Net on the self-supervised task by modifying the final output layer. That is, the output of the network is a $6$-dimension vector which contains the predicted $\hat{a}$, $\hat{b}$ and the corresponding local mean and variance. 

\subsubsection{Loss of Self-Supervised Task}
The loss is inherited from standard regression problem, which minimizes $\mathcal{L}_1$ error between the prediction and ground truth. Given a point cloud with coordinates and triplicate lightness values, the self-supervised pretext task learns a mapping $\hat{Y} = \mathcal{F} (X^{s};\Theta),\hat{Y} = \{\hat{a},\hat{b},\hat{\mu}^a, \hat{\sigma}^a,\hat{\mu}^b, \hat{\sigma}^b\}$, where $\hat{a}$, $\hat{b}$, $\hat{\mu}$ and $\hat{\sigma}$ denote predicted \textbf{a}, \textbf{b} and corresponding local mean and variance, respectively. The loss of self-supervised task can be formulated as:
\begin{equation}
	\mathcal{L}_{ab} = \frac{1}{2N^{s}}\sum ^{N^{s}}_{i=1}(||a_i-\hat{a}_i||_1+||b_i-\hat{b}_i||_1).
\end{equation}

\begin{figure*}[]
	\centering
	\includegraphics[width=15cm]{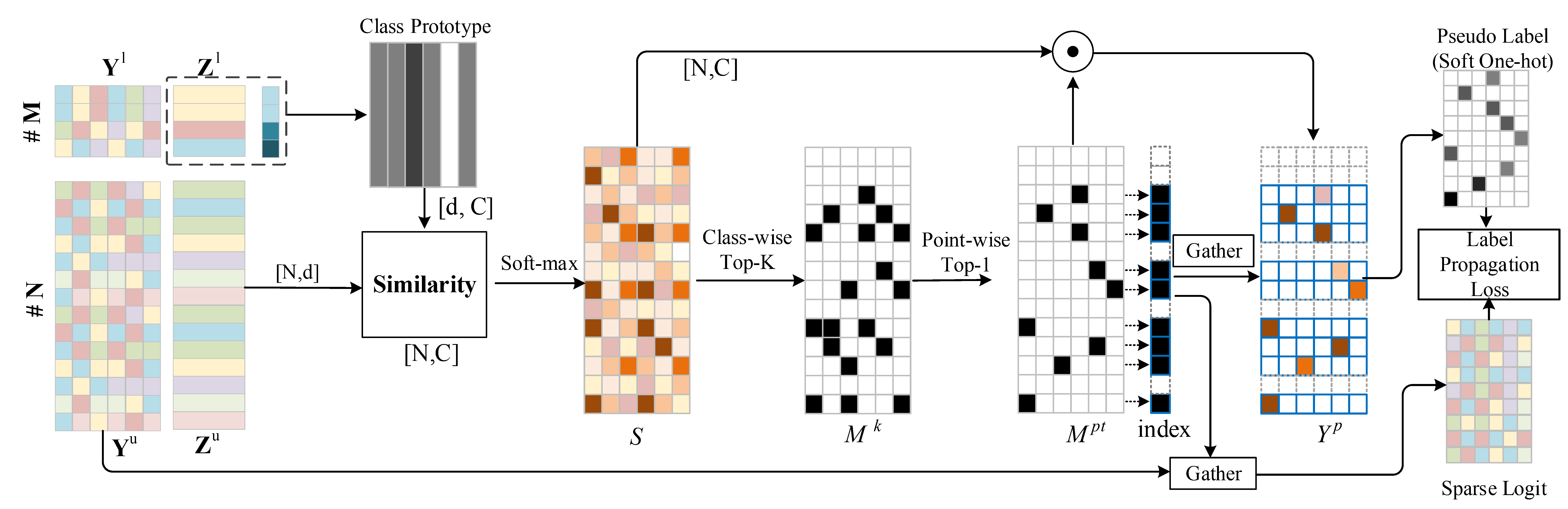}
	\caption{The framework of sparse label propagation. $\odot$ is Hadamard product and \textbf{Gather} denotes the operator of getting elements by index.}
	\label{lp}
\end{figure*}

In addition, to learn the local color distribution of every point, we introduce a local perceptual regularization term. If the network can predict the color distribution (mean and variance) of the neighbors, it can embed the local information which is consistent with the segmentation task using local features for weakly supervised semantic segmentation. Given a point $x_i$ as the centroid, the local neighbor $\mathcal{N}(x_i)$ is calculated by KNN according to the Euclidean distance. The ground truth $\mu_i^{a}$ and $\sigma_i^{a}$ of $a$ channel can be obtained by:
\begin{equation}
	\mu_i^a = \frac{1}{K}\sum _{j=1}^K a_j, \forall {x_j} \in \mathcal{N}(x_i),
\end{equation}
\begin{equation}
	\sigma_i^a = \sqrt{\frac{1}{K}\sum _{j=1}^K (a_j-\mu_i^a)^2 + \varepsilon }, \forall {x_j} \in \mathcal{N}(x_i),
\end{equation}
where $\varepsilon$ is a very small constant. $\mu_i^{b}, \sigma_i^{b}$ can be obtained in the same way.
We formulize the local perceptual regularization term as:
\begin{equation}
	\begin{split}
		\mathcal{L}_{local} = \frac{1}{4N^{s}}\sum ^{N^{s}}_{i=1}(||&\mu_i^a-\hat{\mu_i}^a||_1 + ||\sigma_i^a-\hat{\sigma_i}^a||_1 + \\ ||&\mu_i^b-\hat{\mu_i}^b||_1 + ||\sigma_i^b-\hat{\sigma_i}^b||_1 ).
	\end{split}
\end{equation}

The total loss $\mathcal{L}_{p}$ of self-supervised pretext task can be expressed as:
\begin{equation}
	\mathcal{L}_{p} = \mathcal{L}_{ab} + \mathcal{L}_{local}.
\end{equation}

\subsubsection{Discussion} 

\textit{Why is the knowledge learned from the pretext task beneficial for semantic segmentation?}
\begin{itemize}	
	\item Pretext task learns similar feature distributions as the semantic segmentation task.
	Objects in the same category usually have similar color distribution, for example the vegetation is typically green, and the road is black. The surface color texture of the scene provides ample cues for many categories.
	\item Pretext task embeds the local feature representation.
	We introduce a local perceptual regularization term to constrain the local color distribution to be consistent with the original distribution. Thus, it allows the network to embed more local information. Therefore, it may enhance the embedding of local features for semantic segmentation task.
\end{itemize}

\subsection {Sparse Label Propagation}

Segmentation performance degrades significantly with few labeled points. The main reason is that supervisory information provided by few labeled points can not be propagated well to unlabeled points. Therefore, we use the labeled points to assign pseudo labels for unlabeled points, and further provide additional supervised information to improve the representation of the weakly supervised network.

In order to achieve this goal, the following items require to be taken into account:
1) Computational complexity is not high and memory recourse is not large. Large-scale point clouds usually contain $\sim 10^6$ points, if using all the points as nodes to construct a fully-connected graph, it will consume a lot of memory and computing resources.
2) The anchor points should be sparse. Some ambiguous points should not be given labels to train the network.
3) The propagated label should be soft. The propagated labels should be related to their similarity, and the higher of the similarity, the more similar the label should be.

We design a sparse label propagation method. The overall framework is shown in Figure \ref{lp}. It consists of three parts: class prototype generation, class assignment matrix construction, sparse pseudo-label generation.

\textbf{Class prototype generation.}
In the last two layers of the network, we output embedding $\mathbf{Z}=[z_1^l,z_2^l,...,z_M^l; z_1^u,z_2^u,...,z_N^u] \in \mathcal{R}^{(M+ N)\times d} $ and the corresponding prediction $\mathbf{Y}=[y_1^l,y_2^l,...,y_M^l; y_1^u,y_2^u,...,y_N^u] \in \mathcal{R}^{(M+ N)\times C} $ which comes from $M$ labeled points and $N$ unlabeled points. We use $\mathbf{Z^l}$, $\mathbf{Z^u}$ to represent the embedding of labeled and unlabeled points, respectively. Firstly we generate $C$ prototypes to represent the $C$ classes according to the labeled points. Specifically, we simply take the mean of the labeled point embeddings $\mathbf{Z^l}$ for each class. For class $c$, the prototype $\rho _{c}$ is given by:
\begin{equation}
	\rho _{c}=\frac{1}{\left|\mathcal{I}_{c}\right|} \sum_{z^l_i \in \mathcal{I}_{c}} z^l_i,
\end{equation}
where $\mathcal{I}_{c}$ denotes the embedding sets of class $c$, $c=\{1,2,...,C\}$ for labeled points.

\textbf{Class assignment matrix construction.}
We leverage embedding $\mathbf{Z^u}$ of unlabeled points and the class prototypes to construct a similarity matrix $\mathcal{W} \in \mathcal{R}^{N \times C}$ by:
\begin{equation}
	\mathcal{W} = exp(-\frac{||z_i^u-\rho _{c}||^2}{ \sigma }),
\end{equation}
where $\sigma$ is a hyper-parameter. Each column of $\mathcal{W}$ represents the similarity between the unlabeled point and the class prototypes. We use $soft$-$max$ to convert the similarity into a class assignment probability $S$ as:
\begin{equation}
	S = \mathcal{P}(i \mapsto c) = \frac{exp(\mathcal{W})}{\sum_{c=1}^{C}exp(\mathcal{W})}.
\end{equation}

\textbf{Sparse pseudo label generation.}
There are some points with low similarity to each category. These points are not suitable to provide supervisory information to train the network. Specifically, for each class, according to the class assignment matrix $S$, we select the $top$-$K$ unlabeled points and get the mask $M^{k} \in \{0,1\}^{N\times C}$, where ${m_{ic}^{k}=1}$ means the embedding of $i$-$th$ point is the first $K$ points similar to the class $c$. $N$ is the number of unlabeled points. This is a label expansion method with a balanced number of categories. It can alleviate the category imbalance to a certain extent. 

As an unlabeled point may belong to multiple categories, we choose the most similar category and generate a binary mask. According to $M^{k}$, we get the point mask $M^{pt} \in \{0,1\}^{N}$. $m^{pt}_i =1 $ denotes that the $i_{th}$ point is assigned a pseudo labels. We can get the sparse pseudo label $Y^{p} \in \mathcal{R}^ {N\times C}$ by:
\begin{equation}
	Y^{p} = M^{pt} \odot S,
\end{equation}
where $\odot$ denotes Hadamard product. $Y^{p}$ is the form of soft one-hot. The label propagation loss $\mathcal{L}_{sp}$ can be formulated as:
\begin{equation}
	\mathcal{L}_{sp} = -\frac{1}{||M^{pt}||_1} \sum_{i=1}^N m^{pt}_{i} \sum_{c=1}^C {y}^{p}_{ic} \log y^u_{ic},
\end{equation}
where $y^u_{ic}$ is the probability that the unlabeled point $i$ is classified as category $c$, and ${||\cdot||_1}$ denotes $L1$ norm. 

Compared with the traditional fully-connected graph label propagation method, our method is computation efficient.
The complexity of our method is $\mathcal{O}(NCd)$, while the method of fully-connected graph is $\mathcal{O}((N+M)^2d)$. $C$ is the number of category and $d$ denotes the dimension of $\mathbf{Z^u}$. The magnitude of $C$ is $\sim 10^1$, which is much smaller than $N (\sim 10^6$).

\subsection {Loss of Weakly Supervised Task}
The loss of weakly supervised semantic segmentation includes two terms: segmentation loss and label propagation loss.
\begin{equation}
	\mathcal{L}_{total} = \mathcal{L}_{seg} +  \lambda   \mathcal{L}_{sp}, 
\end{equation}
We utilize a softmax cross-entropy loss on the labeled points. For labeled points, the segmentation loss is formulated as:
\begin{equation}
	\mathcal{L}_{seg} = -\frac{1}{M} \sum_{i=1}^M \sum_{c=1}^C y_{i c} \log \frac{\exp \left(y^l_{ic}\right)}{\sum_{c=1}^C \exp \left(y^l_{ic}\right)},
\end{equation}
where $y_{ic}$ is the ground truth of labeled point $i$. $M$ denotes the number of labeled points.

In the early stage of training, the embedding is unreliable. Class prototypes generated by embedding can not represent classes well. Thus, the label propagation loss should not be introduced to optimize network parameters. As the embedding gets better, the weight of the pseudo loss should increase. Therefore, we introduce a non-linear parameter $\lambda $ to balance the two losses. In this work, $\lambda $ is formulated as:
\begin{equation}
	\lambda  = \left\{
	\begin{aligned}
		&0, & epoch<30 \\
		&e^{\frac{epoch}{max\_epoch}-1},  & otherwise\\
	\end{aligned}
	\right.
\end{equation}
where $epoch$ and $max\_epoch$ denote the current training epoch and the total epochs, respectively.

\section{Experiments and Analysis}
\label{sec:Experiment}
In this section, we firstly introduce the experimental settings. Then, we evaluate the weakly semantic segmentation performance on indoor and outdoor large-scale point clouds, respectively. Furthermore, we perform ablation study to evaluate the importance of the main components.

\subsection{Experiment Settings}

\subsubsection{Dataset Setting for Self-supervised Task}
Previous works on unsupervised 3D representation learning \cite{achlioptas2018learning,gadelha2018multiresolution,hassani2019unsupervised,li2018so,sauder2019self,yang2018foldingnet} mainly focused on ShapeNet \cite{chang2015shapenet} which is a dataset of single-object CAD point cloud models. However, the real-world large-scale point cloud usually contains multiple models of different categories. Pre-training on ShapeNet have poor scalability because of a large domain gap. We choose ScanNet \cite{dai2017scannet} as the pre-training dataset which is a big real-world point cloud dataset containing 2.5 million views in more than 1500 indoor scans. 

To train self-supervised pretext task, we convert the $\textbf{RGB}$ space to the $\textbf{Lab}$ space and split $6$ channels corresponding to the coordinate $(x,y,z)$ and color $(L,a,b)$ into the given channels $(x,y,z,L)$ and the prediction channels $(a,b)$. 

\subsubsection{Dataset Setting for Weakly Supervised Segmentation}

In order to evaluate the performance of our network on weakly semantic segmentation tasks, we experiment on the indoor scene dataset S3DIS \cite{S3DIS} and ScanNetv2 \cite{dai2017scannet} and outdoor dataset Semantic3D \cite{hackel2017isprs}. Note that our method is pre-trained on ScanNet dataset. However, the pre-training task does not use the semantic labels. 

\subsubsection{Implementation Details} 
Here weakly supervised settings are studied. i) 1 point label ($1pt$), we assume there is only one point within each category labeled with ground-truth for each scene. ii) ($x$\%) denote $x$ percentage points with ground-truth. We set $x =\{1,10\}$ in our experiments. iii) The cost of labeling $1\%$ points manually is much higher than $1pt$. We consider a cheap way: super-point ($SPT$), which annotates a local area instead of one point. In all settings, the annotated points are randomly selected.

Our pre-training and weakly semantic segmentation network are built on the efficient backbone of RandLA-Net. The two networks can be trained built upon TensorFlow with a single NVIDIA Titan RTX. We use Adam optimizer with default parameters. The number of neighboring points $K$ is $16$. Both the self-supervised pretext task and weakly semantic segmentation network are trained 80 epochs. We calculate the mean Intersect over Union (mIoU) to evaluate the performance.  

\begin{table}[]
	\centering
	\begin{tabular}{c|c|c|c }
		\toprule     
		Setting & {Method}  & Area5 & 6-fold   \\ \hline
		\multirow{6}{*}{Fully}  
		& PointNet ['17]     & 41   & 47.6 \\ 
		& DGCNN    ['19]     & -    & 56.1 \\  
		& RSNet    ['18]     & 56.5 & - \\ 
		& PointCNN ['18]     & 57.3 & 65.4 \\ 
		& ShellNet ['19]     & -    & 66.8 \\ 
		& RandLA-Net ['20]    & 62.8* & 70.0 \\ \cline{1-4} 
		\multirow{3}{*}{$1pt$ ($0.2\%$) }      
		& $\Pi$ Model ['16] & 44.3  & -\\ 
		& MT          ['17]  & 44.4& -\\ 
		& Xu          ['20] & 44.5 & - \\ 
		\multirow{1}{*}{$ 0.2\%$}   
		& Ours  & 56.4 &-\\ \hline
		\multirow{2}{*}{$1pt$ ($0.03\%$)}   
		& Baseline  & 40.4 & - \\ 
		& Ours  & 45.8 & - \\ 
		\toprule     
		\multirow{1}{*}{ $1\%$}  
		& Ours  & 61.8 & 65.9  \\ \hline
		\multirow{4}{*}{$10\%$}      
		& $\Pi$ Model ['16]  & 46.3   &-\\ 
		& MT          ['17]  & 47.9   &- \\ 
		& Xu          ['20]  & 48.0   &-\\ 
		& Ours                             & 64.0   & 68.1\\ 	
		\bottomrule  
	\end{tabular}
	\caption{Comparisons of performance on S3DIS \cite{S3DIS} (mIoU \%). Area5 denotes evaluation on Area 5. 6-fold is cross evaluation of 6 areas. The superscript * indicates the result evaluated by the official code.}
	\label{s3dis}
\end{table}

\subsection{Evaluation on S3DIS}
We conduct the comparison experiments with state-of-the-art weakly supervised (Weakly) and fully supervised (Fully) on the indoor detaset S3DIS. The former contains: $\Pi$ Model \cite{laine2016temporal}, MT \cite{tarvainen2017mean} and Xu \cite{xu2020weakly}. The latter includes: PointNet \cite{qi2017pointnet}, DGCNN \cite{DGCNN}, RSNet \cite{RSNET}, PointCNN \cite{pointcnn}, ShellNet \cite{shellnet}, RandLA-Net \cite{RandLA-Net}. 

The results are presented in Table \ref{s3dis}. Overall, we observe the consistent improvement in the performance of segmentation with more labeled points. Our $1pt$ denotes only one labeled point for each category in the entire rooms instead of small blocks (\textit{e.g.,} $1\times 1$ meter). Therefore, our $1pt$ represents that total labeled points are about 0.03\%. While the total labeled points of \cite{xu2020weakly} are about 0.2\%. Ours still achieves better performance against Xu \cite{xu2020weakly} at $1pt$ setting. Under $1\%$ setting, we achieve comparable performance against the fully supervised method, RandLA-Net, and even exceed the previously fully supervised methods by a large margin. At $10\%$ setting, ours is overwhelmingly superior to the previous weakly supervised methods and even achieves $1.2\%$ improvement against RandLA-Net \cite{RandLA-Net} on Area 5. These results show the effectiveness of our method.

The qualitative results are shown in Figure \ref{vis_ss} on S3DIS dataset under $1pt$, $1\%$ and $10\%$ settings. It can be seen that at 1\% setting, our method can correctly classify except for challenging boundaries. 
\subsection{Evaluation on ScanNetv2}
We compare our method with five fully supervised methods: PointNet++ \cite{qi2017pointnet++}, SPLATNet \cite{su2018splatnet}, PointCNN \cite{pointcnn}, KPConv \cite{kpconv}, and a weakly supervised method: MPRM \cite{wei2020multi}.
Table \ref{ScanNet} shows the comparison results on test set. From $SPT$ to $10\%$ setting, ours achieves significant gain and make a breakthrough margin compared with fully supervised method PointCNN. At $SPT$ setting, ours gains 7.9\% against MPRM which is trained with subcloud-level semantic labels (subcloud). We cannot directly and quantitatively evaluate the annotated labor of $SPT$ and MPRM. Compared to MPRM, ours needs to draw an extra little region at random. But it reduces the labor of dividing into subclouds and repeatedly annotating semantic category for each subcloud.

\begin{table}[]
	\centering
	\begin{tabular}{c|c| p{0.6cm}}
		\toprule     
		Setting &{ Method}   &  mIoU  \\ \hline
		\multirow{4}{*}{Fully}      
		& PointNet++  ['17]  & 33.9  \\ 
		& SPLATNet ['18]  & 39.3  \\ 
		& PointCNN ['18]  & 45.8 \\ 
		& KPConv   ['19]  & 68.4   \\
		\hline
		\multirow{4}{*}{Weakly}   
		& MPRM (subcloud) ['20]  & 41.1 \\ 	
		& Ours ($SPT$)       & 49.0  \\ 	
		& Ours ($1\%$)       & 51.1  \\ 	
		& Ours ($10\%$)      & 52.0  \\ 	
		\bottomrule  
	\end{tabular}
	\caption{Quantitative results on ScanNetv2 (mIoU \%). }
	\label{ScanNet}
\end{table}

\subsection{Evaluation on Semantic3D}
For outdoor point cloud semantic segmentation, we compare ours against five state-of-the-art fully supervised methods: SnapNet \cite{boulch2017unstructured}, SEGCloud \cite{tchapmi2017segcloud}, ShellNet \cite{shellnet}, KPConv \cite{kpconv} and RandLA-Net \cite{RandLA-Net}. The results of online testing are summarized in Table \ref{Semantic3D}.

Clearly, the more labeled points, the better the semantic segmentation performance is. With 1\% labeled points, we achieve 72.6\% mIoU and outperforms fully supervised ShellNet \cite{shellnet} by the gain of over 3.3\%. With $10\%$ labeled points, our method is close to the current state-of-the-art RandLA-Net. Therefore, our method is also effective in outdoor dataset.
\begin{table}[]
	\centering
	\begin{tabular}{c|c|c| p{0.6cm}}
		\toprule     
		Setting &{Method}  & mIoU & OA  \\ \hline
		\multirow{5}{*}{Fully}      
		& SnapNet  ['17] & 59.1& 88.6  \\ 
		& SEGCloud ['17] & 61.3& 88.1  \\ 
		& ShellNet ['19] & 69.3& 93.2  \\ 
		& KPConv   ['19] & 74.6& 92.9  \\
		& RandLA-Net ['20]& 77.4& 94.8 \\
		\hline
		\multirow{2}{*}{Weakly}   
		& Ours ($1\%$)        & 72.6&	93.7  \\ 	
		& Ours ($10\%$)       & 73.3& 94.0  \\ 	
		\bottomrule  
	\end{tabular}
	\caption{Quantitative results on Semantic3D (reduced-8) \cite{hackel2017isprs}. 'OA' denotes overall accuracy.}
	\label{Semantic3D}
\end{table}

\subsection{Ablation Study}
We conducted ablation studies to evaluate the importance of the main components: self-supervised pre-training task and sparse label propagation. \textit{Scratch} denotes random initialization. \textit{Pre-training} represents that using the parameters of self-supervised pretext task for initialization. $\lambda = 0$ represent without sparse label propagation. $\lambda = 1$ and `nonlinear' denote a constant $1$ and a nonlinear way weighting the sparse label propagation, respectively. 

\textbf{Self-supervised pre-training or training from scratch.}
\begin{table}[]
	\centering
	\begin{tabular}{c c|cc c c}
		\toprule  
		Setting &  &  $1pt$ & $SPT$ & $1\%$ & $10\%$  \\ \hline
		\multirow{1}{*}{Scratch}      
		& $\lambda  = 0$  & 40.4 & 55.8 & 58.6 & 60.6 \\
		\hline
		\multirow{3}{*}{Pre-training}     
		& $\lambda  = 0$  & 45.4& 57.5 & 60.1 & 61.3 \\ 
		& $\lambda  = 1$  & - & - & -& - \\ 
		& nonlinear     & \textbf{45.8}& \textbf{58.2} & \textbf{61.8} & \textbf{64.0} \\ 
		\bottomrule  
	\end{tabular}
	\caption{Comparisons of different components on S3DIS Area 5 \cite{S3DIS} (mIoU \%). }
	\label{SS}
\end{table}

From the comparison of the first and second rows of Table \ref{SS}, the pre-training makes the performance better, while training from scratch drops in performance for 5.0\%, 1.7\%, 1.5\%, 0.7\%. We hypothesize that our pre-training is effective because it can transfer knowledge learned from supervised tasks to subsequent tasks. This encourages the subsequent task to learn a more robust feature distribution. We can still find that the fewer annotation points, the greater the effect of self-supervision. We infer that fewer labeled points cannot learn a good feature representation. Therefore, it needs more additional knowledge to promote feature representation learning.

\textbf{Sparse label propagation.}
We further analyse the effects of sparse label propagation. Comparing the second and fourth rows of Table \ref{SS}, our \textit{nonlinear} obtains 0.4\%, 0.7\%, 1.7\%, and 2.7\% improvement over $\lambda = 0$, respectively. The more labeled points, the better performance of the semantic segmentation achieves. Because more labeled points can learn better representations and produce accurate class prototypes. When $\lambda $ is a constant $1$, the network training collapses. Because the initial training (the first 30 epochs) cannot learn a good feature representation. It is unreasonable to use these features for label propagation. In the process of feature learning gradually getting better, a larger weight is gradually introduced to the loss term. 

\begin{figure}[t]
	\centering
	\subfigure[On S3DIS]{
		\includegraphics[width=3.95cm]{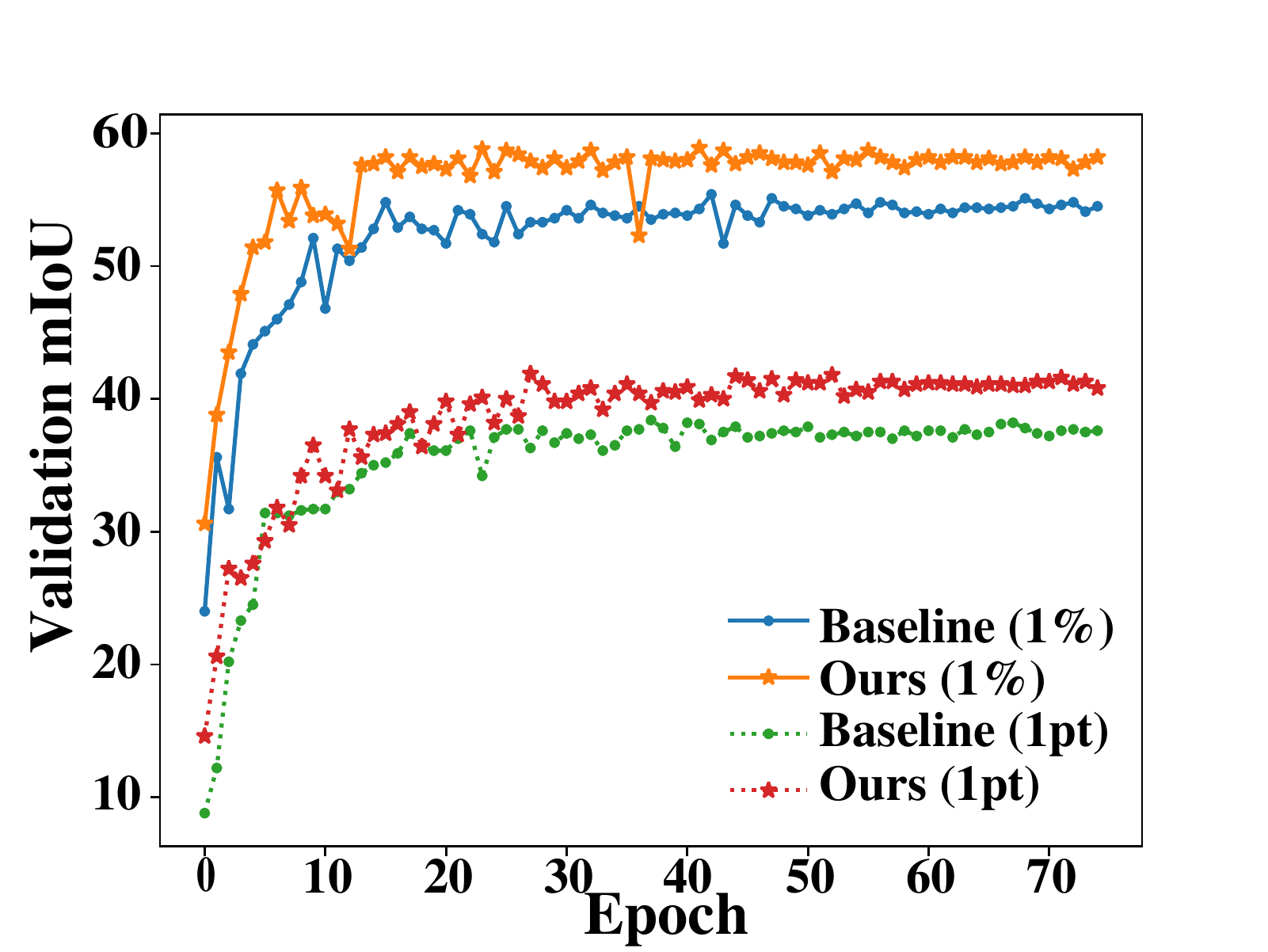}
	}
	\subfigure[On Semantic3D]{
		\includegraphics[width=3.95cm]{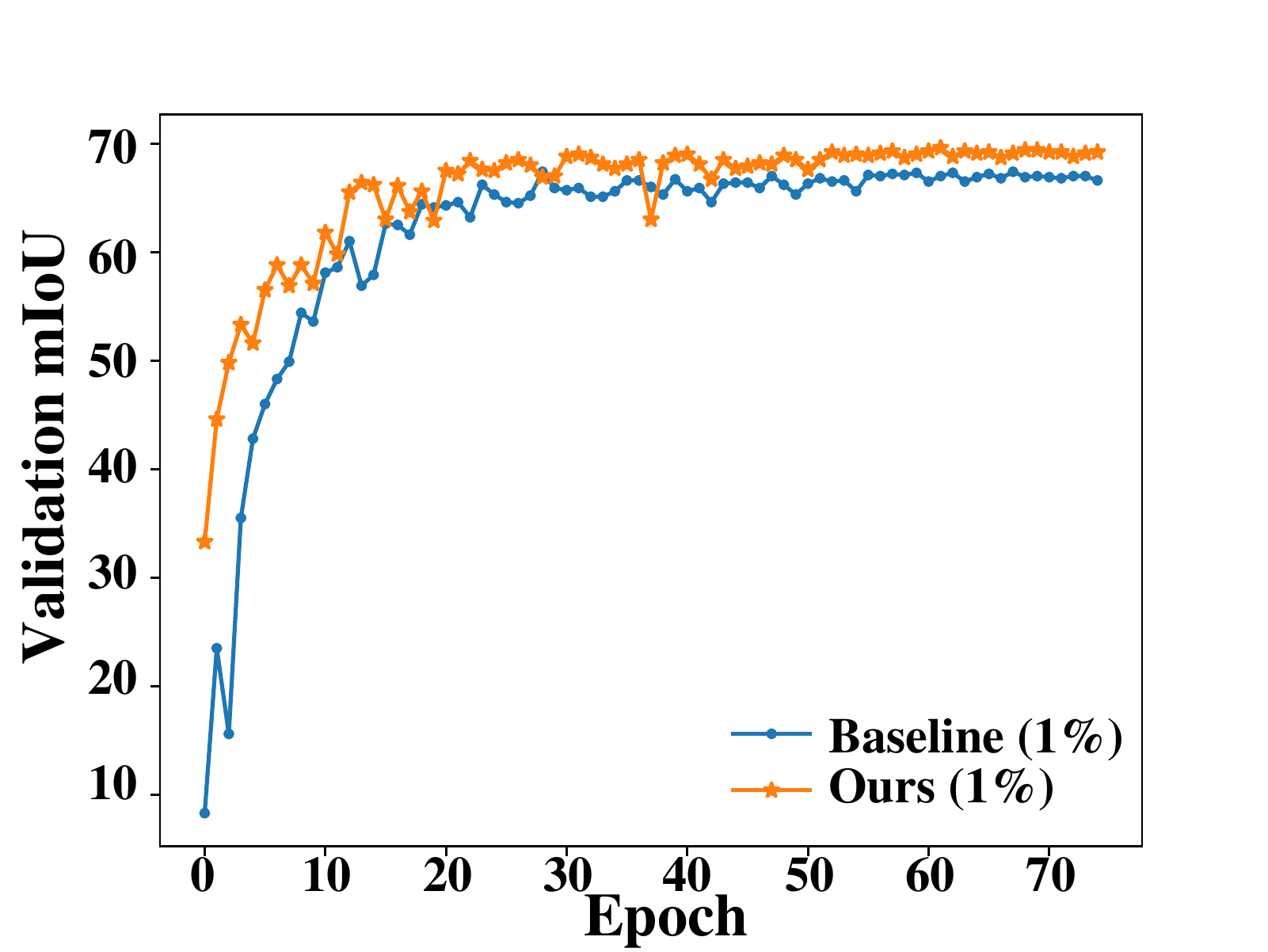}
	}
	\caption{mIoU curve of our method and baseline on the validation set.}
	\label{eff}
\end{figure}

\textbf{Effectiveness and efficiency.}
We conduct a pilot study to understand the effectiveness and show the mIoU curve of our method and baseline on validation set. From Figure \ref{eff}, our method converges faster and has a higher stable value which demonstrates our method is more effective for weakly supervised large-scale point cloud segmentation. 

Like RandLA-Net \cite{RandLA-Net}, we feed the entire scene to the network. We list the training time of each epoch (RandLA-Net: 267s and Ours: 280s) and total test time (RandLA-Net: 115s and Ours: 116s) on Area-5 of S3DIS dataset.
 As the sparse label propagation is introduced in the training phase, training time of per-epoch increases by 13s. But their test time are almost the same. 

\begin{figure}[t]
	\centering
	\includegraphics[width=8.25cm]{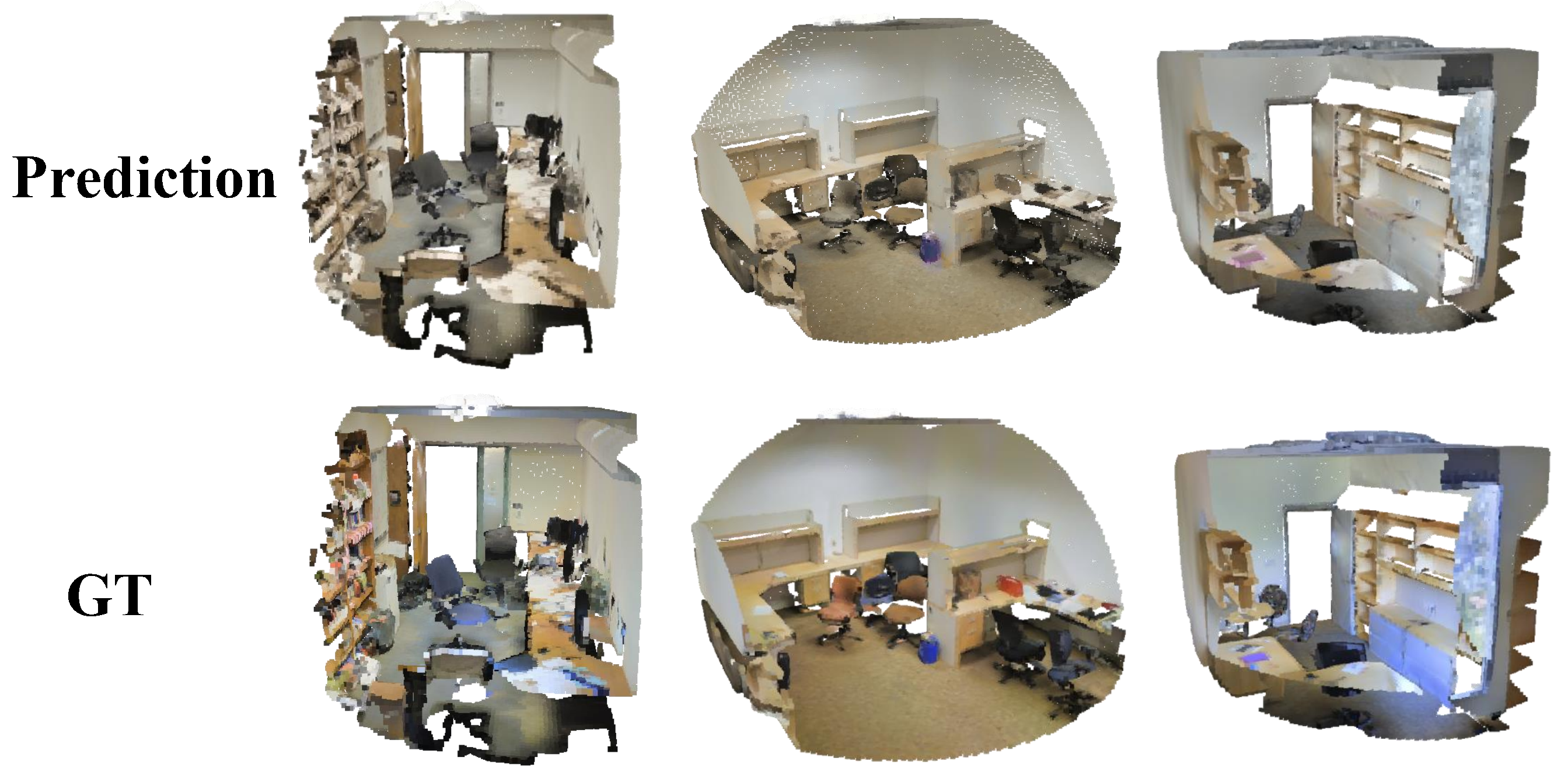}
	\caption{Visualize the results of point cloud colorization.}
	\label{vis}
\end{figure}

We also visualize the results of colorization in Figure \ref{vis} on S3DIS. Although our point cloud colorization focuses on the representation learning, it can be seen that the colorization is close to real-world color.

\section{Conclusion}
We present a weakly supervised semantic segmentation method for large-scale point cloud. With the help of a self-supervised pretext task and sparse label propagation, our method significantly outperforms the weakly supervised and almost reaches the accuracy of fully-supervised methods on three challenging large-scale point cloud dataset. The experimental results show self-supervised knowledge transfer is an effective way to improve weak supervised performance and our method make a breakthrough margin against the compared weakly supervised method. Furthermore, the more the labeled points, the better performance our method achieves, which is consistent with our expectation. In the future, we expect future work to explore more effective self-supervised knowledge representation for weakly or semi-supervised point cloud tasks.

\section{Acknowledgments}
This work is supported by the National Natural Science Foundation of China under Grant 61876161, Grant 61772524, the National Key Research and Development Program of China No.2020AAA0108301, and Natural Science Foundation of Shanghai No.20ZR1417700. CAAl-Huawei Mind-Spore Open Fund.

\bibliography{aaai}

\end{document}